\documentclass[letterpaper, 10 pt, conference]{ieeeconf}  

\IEEEoverridecommandlockouts                              

\usepackage{graphics} 
\usepackage{graphicx}
\usepackage{url}
\usepackage{amsmath}
\usepackage{amsfonts}
\usepackage{xcolor}
\usepackage{listings}
\usepackage{subcaption}

\title{\LARGE \bf
    ART/ATK: A research platform for assessing and mitigating the sim-to-real gap in robotics and autonomous vehicle engineering
}

\author{Asher Elmquist$^{1}$, Aaron Young$^{1}$, Thomas Hansen$^1$, 
	Sriram Ashokkumar$^{1}$, Stefan Caldararu$^{1}$, Abhiraj Dashora$^{1}$, \\
	Ishaan Mahajan$^{1}$, Harry Zhang$^1$, Luning Fang$^1$, He Shen$^{2}$, Xiangru Xu$^{3}$, Radu Serban$^{1}$, and Dan Negrut$^{1}$ 
\thanks{$^{1}$Simulation-Based Engineering Lab, Department of Mechanical Engineering, University of Wisconsin-Madison, Madison WI, USA} %
\thanks{$^{2}$Robotics Lab, Department of Mechanical Engineering, California State University, Los Angeles, USA
	{\tt\small hshen6@calstatela.edu}}%
\thanks{$^{3}$Autonomous \& Resilient Controls Lab, Department of Mechanical Engineering, University of Wisconsin-Madison, WI, USA
        {\tt\small xiangru.xu@wisc.edu}}%
}

\include{macros}

\begin{document}

\maketitle
\thispagestyle{empty}
\pagestyle{empty}

\begin{abstract}	
	We discuss a platform that has both software and hardware components, and whose purpose is to support research into characterizing and mitigating the sim-to-real gap in robotics and vehicle autonomy engineering. The software is operating-system independent and has three main components: a simulation engine called Chrono, which supports high-fidelity vehicle and sensor simulation; an autonomy stack for algorithm design and testing; and a development environment that supports visualization and hardware-in-the-loop experimentation. The accompanying hardware platform is a 1/6th scale vehicle augmented with reconfigurable mountings for computing, sensing, and tracking. Since this vehicle platform has a digital twin within the simulation environment, one can test the same autonomy perception, state estimation, or controls algorithms, as well as the processors they run on, in both simulation and reality. A demonstration is provided to show the utilization of this platform for autonomy research. Future work will concentrate on augmenting ART/ATK with support for a full-sized Chevy Bolt EUV, which will be made available to this group in the immediate future.
\end{abstract}

\section{INTRODUCTION}
\label{sec:intro}

Simulation has the potential to facilitate the development of algorithms that improve the autonomy of self-driving vehicles and robotic systems \cite{PNASsimRobotics2021}. The value proposition is tied to the cost-effective manner in which data can be generated in simulation as well as to the ease and safety with which candidate solutions can be tested and iterated upon. Using simulation in robotics requires a dynamics engine and a model \cite{karenSimRobotics2020}. 
Should one make the necessary investment to understand both how the simulator and model should be configured, simulation provides insights that are difficult to obtain in physical testing, e.g., complete state information and quantitative insights about the interaction between the robot and environment it operates in. Unfortunately, these insights do not always lead to decisions that work well in reality due to the simulation-to-reality gap \cite{sim2realGapEssex1995}. How to close this sim-to-real gap remains an open problem, and solutions have been proposed that include randomizing the experience of the robot inside the simulator \cite{domainRandomizationAbbeel2017,sim2Real2018}, using adversarial learning to capture unknown components of the model via ghost external perturbations in an adversarial reinforcement learning framework \cite{pintoAdversRL2017}, using ensembles of models \cite{levineEPOpt2016}, using a mix of simulation-generated and real-world data to train robots \cite{farchyStone2013,foxSim2RealClosing2019}, etc.

Despite these and other similarly valuable contributions, the community lacks an objective understanding of what exactly produces the sim-to-real gap. An open source autonomy research testbed can be a catalyst for research in understanding and mitigating the sim-to-real gap. Together with its digital twin in a simulated environment, they can be used for improvement and development of algorithms in autonomy research. Some projects have been developed along this line. For example, the open-source race-car project MuSHR \cite{srinivasa2019mushr}. MuSHR is ROS-anchored and uses Gazebo for simulation \cite{gazebo}. The hardware platform is a 1/10th form-factor vehicle, has one 2D lidar sensor, one stereo camera, and a Jetson Nano processor. 
Another well known platform is MIT RACECAR \cite{racecarMIT2022,racecarMIT-github2022}. The vehicle has a form factor of 1/10th, and is equipped with a stereo camera and a 2D lidar. The software stack, which is ROS-anchored, is provided as a Docker image. Simulation support comes via Gazebo. 
The Cat \cite{catPlatform2022} also has a 1/10th form factor and it uses two IR sensors to chase an IR LED mounted on a remote control car. The simulator is created using MATLAB. 
Donkey \cite{donkeyCar2022} is a miniaturized hobbyist racing vehicle equipped with Raspberry Pi 3b+ and a wide-angle camera. Donkeycar uses Python for controlling the car, and Unity \cite{unityGaming} with PhysX \cite{physxNVIDIA} for simulation. 

However, these platforms share some limitations, e.g., 1) difficult to expand beyond the current vehicles, adding more sensors, etc.; and 2) limited capability in simulating the dynamics and robot--environment two-way interactions. To address this, we developed an open-source platform for autonomy research that allows users to easily swap or add sensors, quickly experiment autonomy algorithms with a powerful and accurate physics based simulation engine, and seamlessly transfer the simulated experience to reality. Our contribution is twofold. First, we established a companion software and hardware platform that in tandem facilitate two types of technical pursuits: research into algorithms for autonomy in mobility, in the context of wheeled and tracked vehicles; and quantitative characterization of the sim-to-real gap in robotics. Second, we developed an autonomy toolkit (ATK) which is a set of command line tools for building the container system that underpins ART. 

The rest of this contribution is organized as follows. In Section \ref{sec:system}, the overview of the system design is presented. Section \ref{sec:atk} highlights the autonomy toolkit framework used to develop the software and hardware platform. Section \ref{sec:art} describes the autonomy research testbed by outlining the autonomy stack and its structure, the bridge to the Chrono simulator, the 1/6th vehicle platform, and its digital twin. Section \ref{sec:demo} shows a demonstration. Closing remarks and directions of future work are provided in Section \ref{sec:conclusion}.

\section{SYSTEM OVERVIEW}
\label{sec:system}

\begin{figure}[t]
	\centering
	\includegraphics[width=0.8\linewidth]{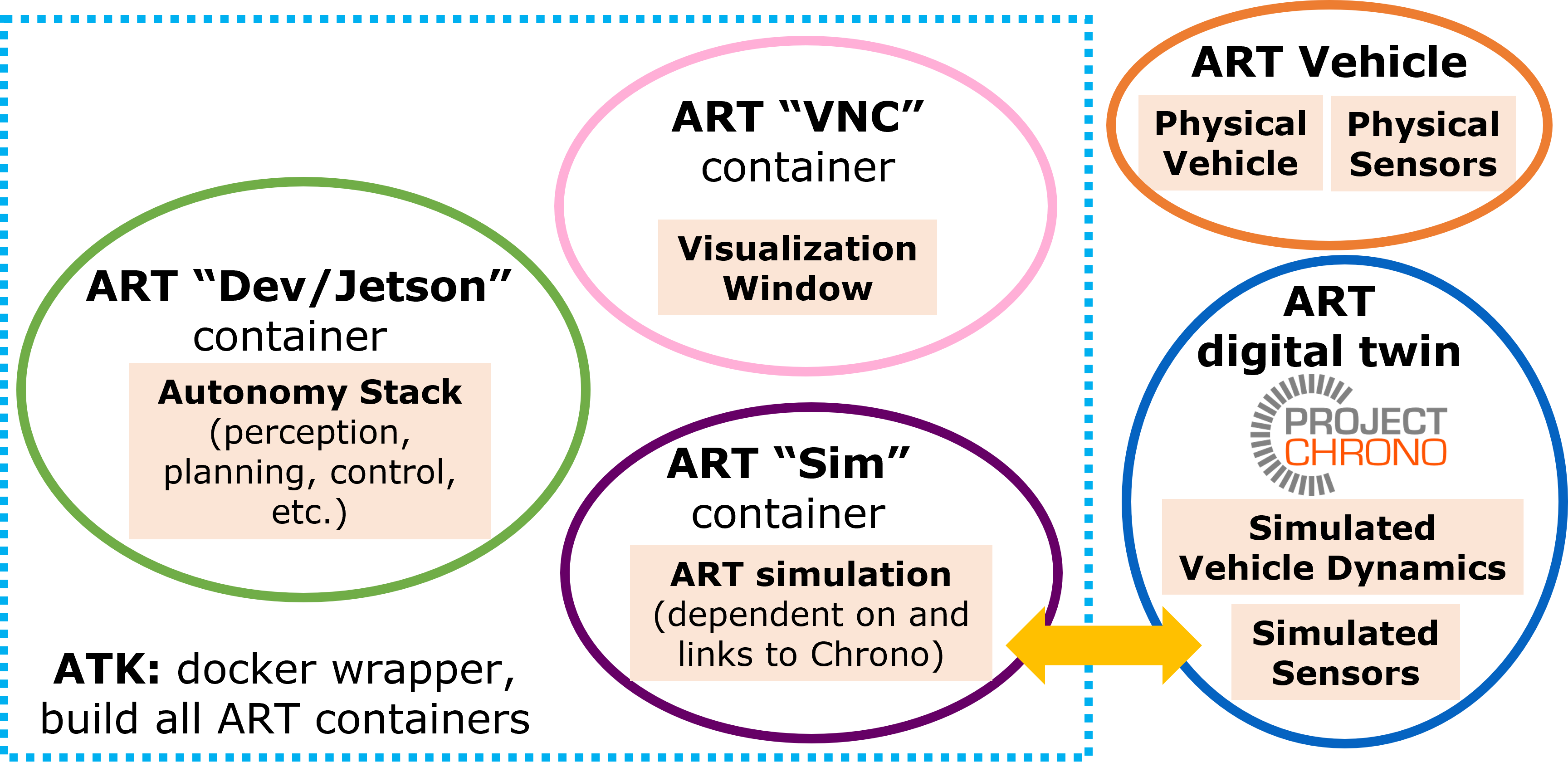}
	\caption{Overall design of the software and hardware platform.}
	\label{fig:system_overview}
\end{figure}

\begin{figure}[t]
	\centering
	\includegraphics[width=0.5\linewidth]{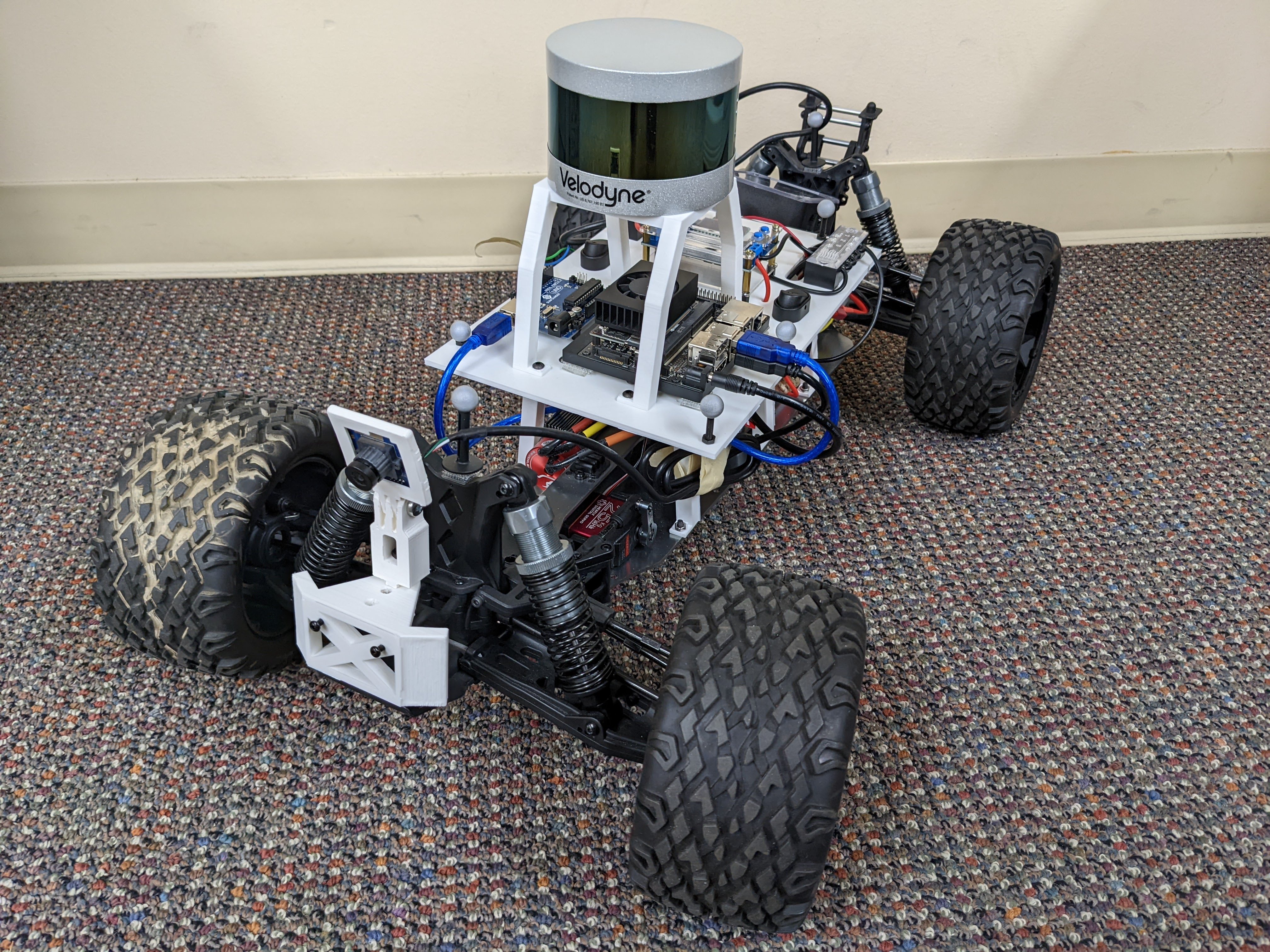}
	\caption{Fully equipped ART vehicle platform}
	\label{fig:art_vehicle}
\end{figure}

As shown in Fig.~\ref{fig:system_overview}, the proposed platform in this work consist of a software environment development tool - Autonomy Toolkit (ATK) and an Autonomous Research Testbed (ART, consist of the hardware system and its digital twin). ATK provides the tools to build the environment and collect, visualize, and analyze data coming from ART. ART consists of a hardware platform and its digital twin in simulation. The hardware and software components of ART shares the autonomy stack, which assembles all algorithms (e.g., perception, state estimation, planning, and control) that enable the intelligence of the robot in both simulation and reality. The hardware platform uses a 1/6th scale vehicle shown in Fig.~\ref{fig:art_vehicle}. Its suspension and steering mechanism closely mirror those of a full-size vehicle, providing a payload capacity for the same class of sensors for full-sized vehicles in off-road conditions. The digital twin is created using the open-source simulation engine Chrono \cite{chronoOverview2016,pyChronoCondaWebSite}, which provides high fidelity simulation of vehicle dynamics \cite{ChronoVehicle2019}, vehicle-terrain interactions \cite{chronoSCM2019,weiTracCtrl2022}, sensors \cite{asherSensorSimulation2021}, and support for multi-vehicle simulation \cite{synchrono2020}. The software is containerized to provide an OS agnostic platform. Setting up an autonomy stack is facilitated by providing ROS bridges and interfaces to other system components to enable researchers to quickly compare and evaluate algorithms associated with the autonomy stack.

\section{AUTONOMY TOOLKIT}
\label{sec:atk}


ATK provides a modular and portable framework for developing, testing, and deploying autonomous algorithms in simulation and reality \cite{atk-pypi}. It is a Python package that leverages Docker by recruiting Docker Compose to build and deploy a multi-container system. ATK is separated into ``services'' that can be combined to produce a container network that supports complex interactions across a variety of projects. The development framework is built into two main components: the ATK Python package and a container system.

\subsection{ATK Python Package}
\label{section:autonomy-toolkit-package}

The ATK Python package leverages the multi-container architecture to facilitate development of autonomous systems with user-defined workflows. Here, YAML configuration files are combined with defaults into a configuration file readable by Docker Compose. From there, ATK makes calls to Docker Compose based on the resulting output. ATK was built with modularity and expandability in mind, and its functionality is agnostic of platforms that it's deployed on. Individual hardware platforms and control stacks may implement their own containers and customize the default configurations \cite{atk-art2022}. ATK can also be used to generate custom containers outside the original scope of the toolkit. 
	
\subsection{Container System}
\label{section:container-system}

Distributed with the ATK package are many predefined and customizable utilities that are used to generate the images and containers. The two primary services are \texttt{dev} and \texttt{vnc}. 

	\textbf{\texttt{dev}:} This is the primary component that is used for algorithm development. It is defined by a custom Dockerfile which utilizes build arguments specified through a configuration file to generate an image specific to the project being developed. By default, \texttt{dev} builds on the ROS2 Galactic distribution, and is assumed to be the primary container that initializes a shell environment when launched. The directory that holds the configuration file is mounted into the container for data persistence upon container termination. Other utilities and configuration are preformed to further enhance the generality of the toolkit and remove OS dependence.

	\textbf{\texttt{vnc}:} A cross-platform visualization system, Virtual Network Computing (VNC), is built upon for on-line visualization purposes. To use VNC visualization, \texttt{dev} (or any other container run using ATK) must simply attach to the same network and configure its \texttt{DISPLAY} variable to that of the \texttt{vnc} hostname. NoVNC, a browser based VNC client, can then be used to view any displayed windows from \texttt{dev} from any internet browser \cite{atk-art2022}. This setup facilitates interactions among distributed containers and supports hardware-in-the-loop experiments, where the simulation container is deployed on separate hardware from the autonomy stack.

\begin{figure}[t]
	\centering
	\includegraphics[width=0.8\linewidth]{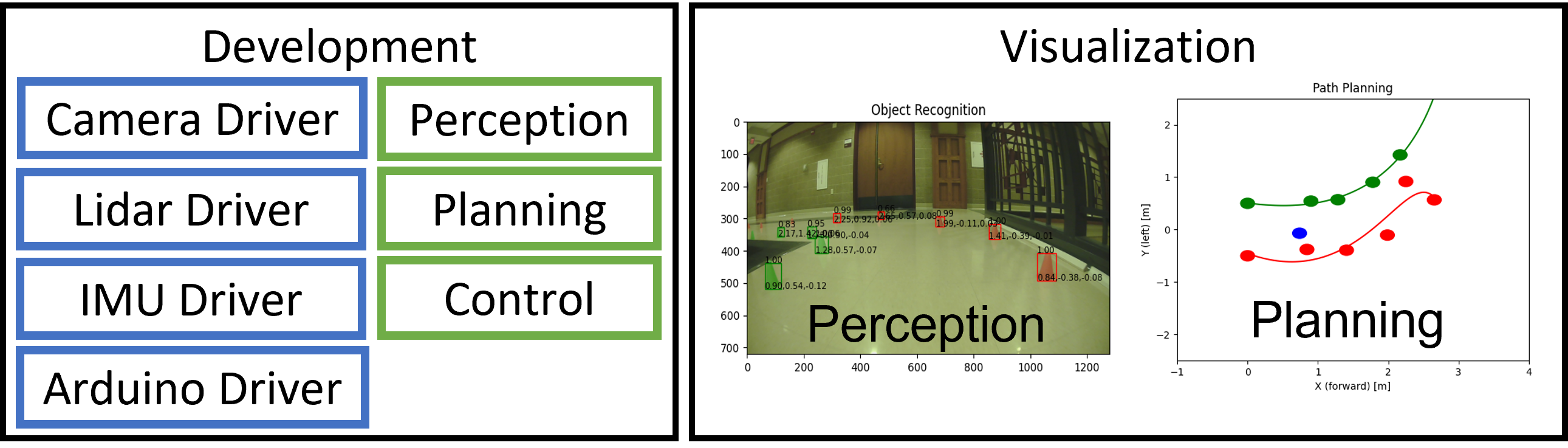}
	\caption{Environment and setup of the ART physical vehicle.}
	\label{fig:dev_env_real}
\end{figure}
\begin{figure}[t]
	\centering
	\includegraphics[width=0.8\linewidth]{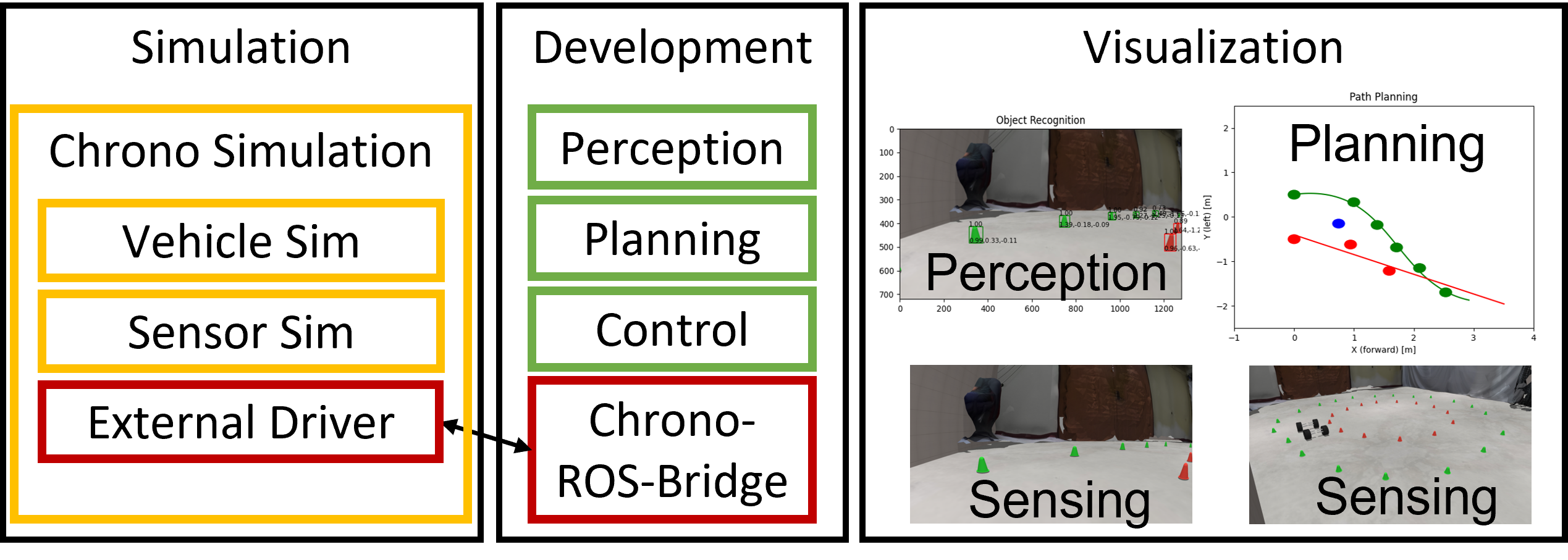}
	\caption{Environment and setup of the ART simulation.}
	\label{fig:dev_env_sim}
\end{figure}

Since the environment is containerized, it can be deployed onto a physical vehicle, with optional hardware-specific optimizations. An example of the containerized system can be seen in Figs. \ref{fig:dev_env_real} and \ref{fig:dev_env_sim}, which illustrate the services when running on the real vehicle vs. simulation, respectively. 

\section{AUTONOMY RESEARCH TESTBED}
\label{sec:art}

\subsection{Simulation}
\label{sec:simulation}

To use simulation for designing and testing autonomous algorithms, a bridge to Chrono was developed to allow direct integration within the containerized system for ART. The bridge, along with the custom Chrono container, is an example of the modular environment produced by ATK. The conduit passes back and forth JSON messages so users can send \textit{any} data between ROS and Chrono. Generic publishers and subscribers (a feature of ROS2) are leveraged to allow for user defined message types and topic names to be used. The package, written in C++, is open source and available under a BSD3 license. 

On the Chrono side, the bridge builds directly on core functionality available through the utility class ChSocket. Additional hooks were provided that wrap the JSON generation from user code. This was done to facilitate Python wrapping with PyChrono, the Python bindings to the C++ Chrono API. This allows the researchers to leverage the rapid development process enabled by Python. Additionally, to allow custom message types to be sent between ROS and Chrono, custom functors may be implemented to generate and/or parse custom message formats.

\subsection{Autonomy Stack}
The autonomy stack developed for ART is built on top of ROS2 and includes basic implementations of autonomy algorithms to enable autonomy experiments in simulation and reality. The stack is basic; there is nothing particularly novel in its implementation. It utilizes publicly available algorithms and packages typically shipped with ROS2. This is because the focus of ART is not on the autonomy stack itself, but the extent to which the algorithms are transferable between sim and reality. Against this backdrop, the perception algorithm is a custom trained instance of Faster-RCNN \cite{ren2016faster} built on a MobileNetV3 \cite{mobileNetV3} network. MobileNet is intended for mobile phone CPUs, but was adopted for ART considering the limited compute power available on the vehicle. For the example used later in this contribution, in which a vehicle navigates down a lane set up with cones, the perception algorithm was trained using both simulated and real images. Based on the bounding box output from Faster-RCNN, 3D cone positions are generated at each time step and a simple planner and controller are used to keep the vehicle on the path defined by the cones. To control the car, throttle, braking, and steering inputs are communicated to the car via an Arduino processor. ROS2 stock interface packages are utilized to communicate with sensors.

\subsection{Vehicle Platform}
The vehicle platform is modified from a 1/6th scale remote controlled car. With a 47 cm wheel base and a 34 cm track width, the vehicle is large enough to carry common used sensors for autonomy research. The base vehicle includes a double wishbone independent suspension at the front and rear. It uses a 1300 KV brushless motor for driving and a 25 kg-cm servo for Pitman arm steering. The motor and servo are controlled by an Arduino micro-controller. The vehicle is also equipped with an onboard computer Jetson Xavier NX, an IMU, a USB camera and a VLP-16 lidar. The base plate and all mounting components are designed for easy manufacturing though laser cut or 3D printing. The final setup is shown in Fig. \ref{fig:art_vehicle} with the electronics mounted above the motor and ESC on the base RC car. The lidar is lifted to be clear of the rest of the vehicle, and the camera is mounted to the front bumper. In addition to electronics and sensors, motion capture tracking targets can be mounted to the car using a scattering of holes across the vehicle.

\subsection{Vehicle Digital Twin}
\begin{figure}[ht]
	\centering
	\begin{subfigure}{.49\linewidth}
		\centering
		\includegraphics[width=\linewidth]{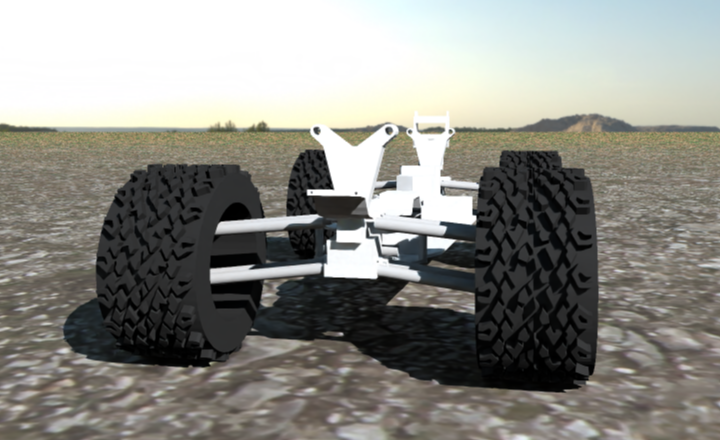}
		\caption{Front view of ART model.}
		\label{fig:chrono_model_front}
	\end{subfigure}%
	\hspace{.01cm}
	\begin{subfigure}{.49\linewidth}
		\centering
		\includegraphics[width=\linewidth]{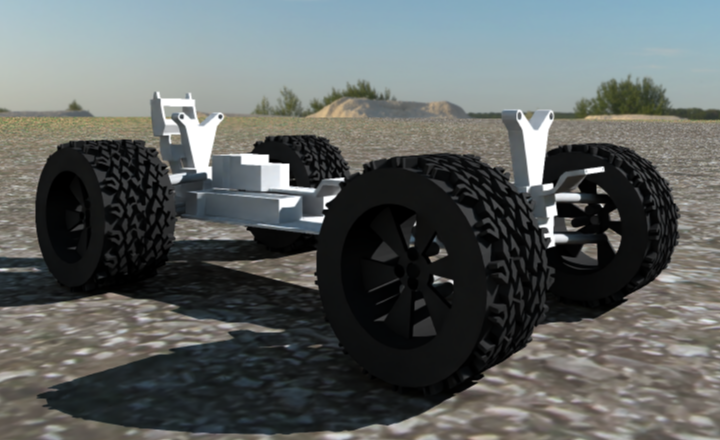}
		\caption{ISO view of ART model.}
		\label{fig:chrono_model_iso}
	\end{subfigure}
	\caption{Chrono::Vehicle model of ART as visualized with Chrono::Sensor.}
	\label{fig:chrono_model}
\end{figure}

The digital twin of the vehicle is modeled in Chrono using Chrono::Vehicle \cite{ChronoVehicle2019} and Chrono::Sensor \cite{asherSensorSimulation2021}. The simulated model is implemented with the same double wishbone suspension and a linear spring-damper as those on the real vehicle. The Chrono model of the car is rendered in Fig. \ref{fig:chrono_model} using Chrono::Sensor and highlights the mesh representation of the car along with the double wishbone configuration.

The steering model uses the Chrono::Vehicle Pitman arm template, allowing actuation of the steering arm. The maximum steering angle was calibrated from a set of minimum radius turn tests using motion tracking. The motor model is a simple linear torque-speed curve, with decreasing power with motor speed. 

\section{DEMONSTRATION}
\label{sec:demo}

\begin{figure}[ht]
	\centering
	\begin{subfigure}{\linewidth}
		\centering
		\includegraphics[width=0.8\linewidth]{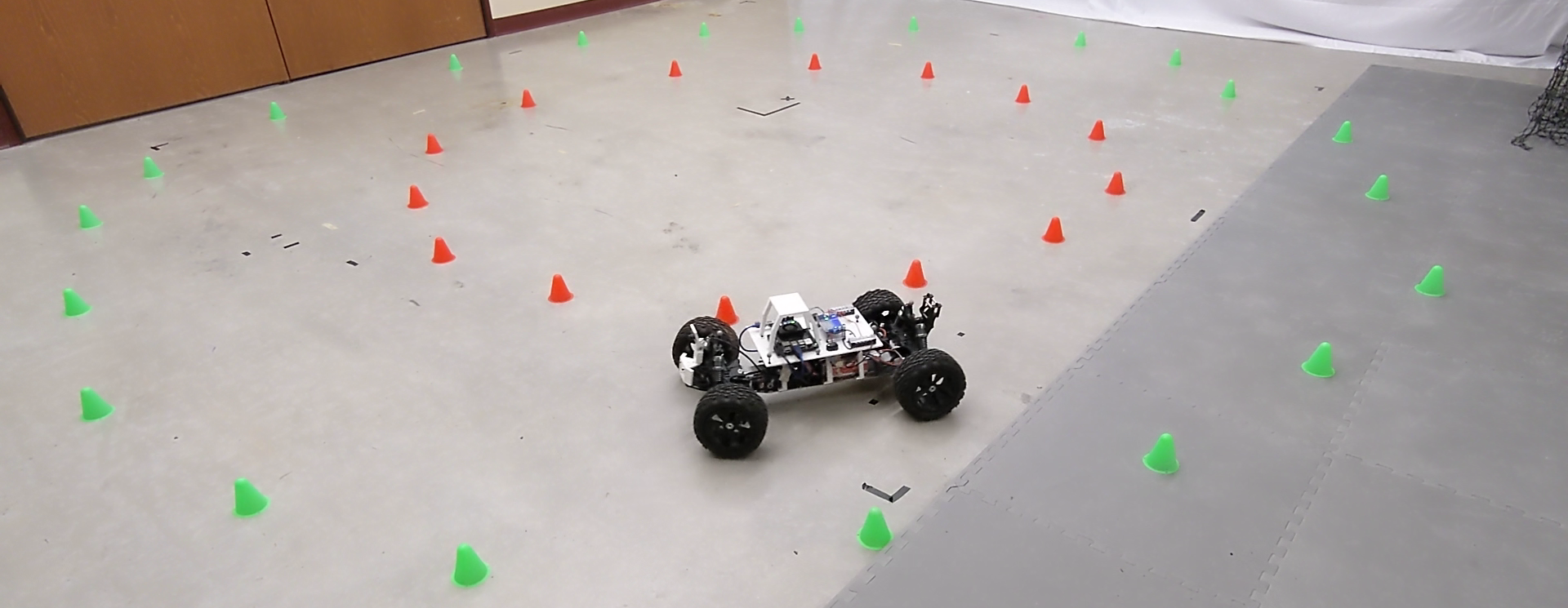}
		\caption{Real setup}
		\label{fig:real_3rdperson}
	\end{subfigure}
	\begin{subfigure}{\linewidth}
		\centering
		\includegraphics[width=0.8\linewidth]{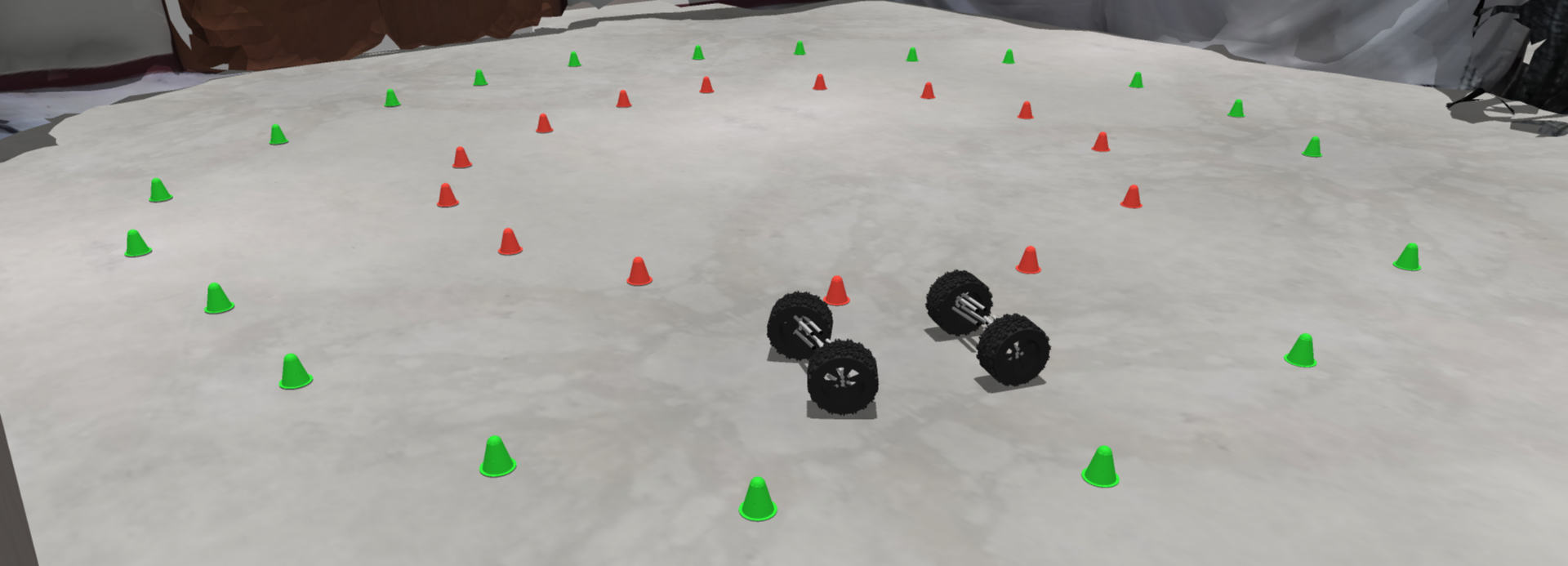}
		\caption{Simulated setup}
		\label{fig:sim_3rdperson}
	\end{subfigure}
	\caption{Real and simulated scenarios with same cone locations.}
	\label{fig:sim_real_1_to_1}
\end{figure}

To demonstrate ART at work, a course defined by cones was setup in our motion tracking lab. The cone locations were recorded and injected into a simulation of the same driving scenario. The vehicle then drove along identical paths in simulation and reality. Images of this setup from a similar location for real and sim are provided in Fig. \ref{fig:sim_real_1_to_1}, which shows the same cone paths and the vehicle in a similar location on the path. To navigate, the vehicle exclusively uses its front facing camera. This demonstration is included in the supporting video and can be found online \cite{art-iros-video}.

\begin{figure}[ht]
	\centering
	\begin{subfigure}{\linewidth}
		\centering
		\includegraphics[width=0.8\linewidth]{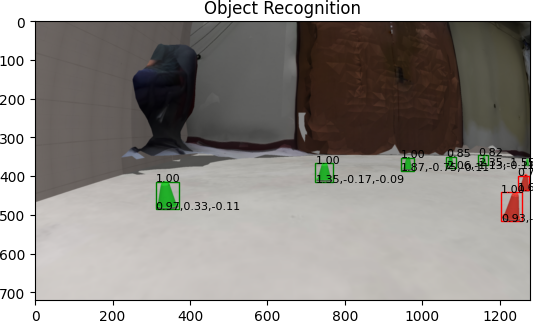}
		\caption{Perception on a simulated image}
	\end{subfigure}
	\begin{subfigure}{\linewidth}
		\centering
		\includegraphics[width=0.8\linewidth]{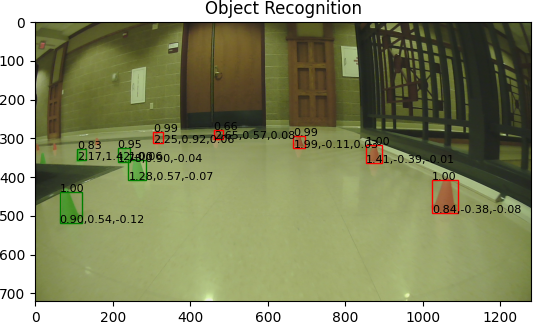}
		\caption{Perception on a real image}
	\end{subfigure}%
	\caption{Comparison of perception in simulation and reality.}
	\label{fig:autonomy_stack_perception}
\end{figure}

\begin{figure}[ht]
	\centering
	\begin{subfigure}{\linewidth}
		\centering
		\includegraphics[width=0.8\linewidth]{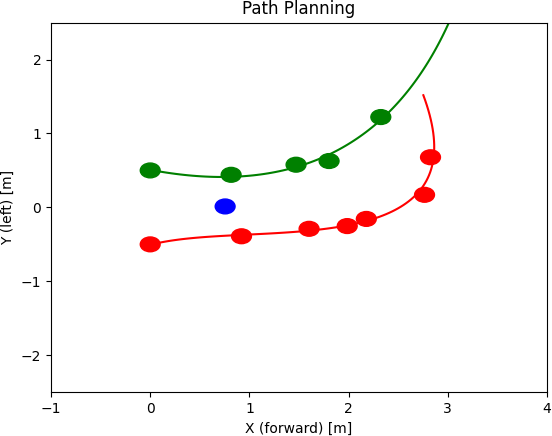}
	\end{subfigure}
	\caption{Path planning with cone detection, boundary estimation, and the target point determination.}
	\label{fig:autonomy_stack_planning}
\end{figure}

To further demonstrate the algorithms used by ART, we show examples of the intermediate output from the perception and planning stages. Perception results from simulation and reality are show in Fig. \ref{fig:autonomy_stack_perception}, where the detected bounding boxes, classes, confidence, and estimated 3D location relative to the vehicle are overlaid on the image. The results from the planning stage are shown in Fig. \ref{fig:autonomy_stack_planning}, where the detected cones in 2D (circles), estimated curve of the boundary, and the determined target point for control (blue) are labeled.

\section{CONCLUSION}
\label{sec:conclusion}

This brief outlines an open-source autonomy research testbed (ART) whose purpose is twofold: conduct research in autonomy for wheeled/tracked vehicles, in on/off-road conditions; and, investigate the sim-to-real gap in robotics -- understand what causes it, and how it can be controlled. Looking ahead, we will equip ART with additional sensors (e.g. GPS, IMU, and extra cameras including a stereo camera). Given that we can track the position of the vehicle in reality with millimeter accuracy via a motion capture system, this richer family of sensors will allow us to investigate more expeditiously better sensor models, as well as perception, state estimation, planning, and controls algorithms, and understand how inaccuracy in different components of the autonomy stack propagate downstream and cause the sim-to-real gap. Another future direction is tied to setting up ART to work with tracked vehicles and in off-road conditions.

Finally, ATK can be used to facilitate autonomy algorithm development by allowing researchers to configure a custom set of containers to host their development, deployment, simulation, and visualization needs. For instance, this platform is assisting UW-Madison students in their SAE-sponsored AutoDrive Challenge II.





\section*{ACKNOWLEDGMENT}
This work was carried out in part with support from National Science Foundation projects CPS1739869, CISE1835674, and OAC2209791. Special thanks to the Society of Automotive Engineers (SAE) and General Motors for their support through the AutoDrive Challenge Series II competition.


\bibliographystyle{ieeetr}

\begin{thebibliography}{10}

\bibitem{PNASsimRobotics2021}
H.~Choi, C.~Crump, C.~Duriez, A.~Elmquist, G.~Hager, D.~Han, F.~Hearl,
  J.~Hodgins, A.~Jain, F.~Leve, C.~Li, F.~Meier, D.~Negrut, L.~Righetti,
  A.~Rodriguez, J.~Tan, and J.~Trinkle, ``On the use of simulation in robotics:
  Opportunities, challenges, and suggestions for moving forward,'' {\em
  {Proceedings of the National Academy of Sciences}}, vol.~118, no.~1, 2021.

\bibitem{karenSimRobotics2020}
K.~Liu and D.~Negrut, ``The role of physics-based simulators in robotics,''
  {\em Annual Review of Control, Robotics, and Autonomous Systems - accepted},
  2020.

\bibitem{sim2realGapEssex1995}
N.~Jakobi, P.~Husbands, and I.~Harvey, ``Noise and the reality gap: The use of
  simulation in evolutionary robotics,'' in {\em European Conference on
  Artificial Life}, pp.~704--720, Springer, 1995.

\bibitem{domainRandomizationAbbeel2017}
J.~Tobin, R.~Fong, A.~Ray, J.~Schneider, W.~Zaremba, and P.~Abbeel, ``Domain
  randomization for transferring deep neural networks from simulation to the
  real world,'' in {\em 2017 IEEE/RSJ international conference on intelligent
  robots and systems (IROS)}, pp.~23--30, IEEE, 2017.

\bibitem{sim2Real2018}
X.~B. Peng, M.~Andrychowicz, W.~Zaremba, and P.~Abbeel, ``Sim-to-real transfer
  of robotic control with dynamics randomization,'' in {\em 2018 IEEE
  International Conference on Robotics and Automation (ICRA)}, pp.~1--8, May
  2018.

\bibitem{pintoAdversRL2017}
L.~Pinto, J.~Davidson, R.~Sukthankar, and A.~Gupta, ``Robust adversarial
  reinforcement learning,'' in {\em International Conference on Machine
  Learning}, pp.~2817--2826, PMLR, 2017.

\bibitem{levineEPOpt2016}
A.~Rajeswaran, S.~Ghotra, B.~Ravindran, and S.~Levine, ``{EPO}pt: Learning
  robust neural network policies using model ensembles,'' 2016.

\bibitem{farchyStone2013}
A.~Farchy, S.~Barrett, P.~MacAlpine, and P.~Stone, ``Humanoid robots learning
  to walk faster: From the real world to simulation and back,'' in {\em
  Proceedings of the 2013 international conference on Autonomous agents and
  multi-agent systems}, pp.~39--46, 2013.

\bibitem{foxSim2RealClosing2019}
Y.~Chebotar, A.~Handa, V.~Makoviychuk, M.~Macklin, J.~Issac, N.~Ratliff, and
  D.~Fox, ``Closing the sim-to-real loop: Adapting simulation randomization
  with real world experience,'' in {\em 2019 International Conference on
  Robotics and Automation (ICRA)}, pp.~8973--8979, IEEE, 2019.

\bibitem{srinivasa2019mushr}
S.~S. Srinivasa, P.~Lancaster, J.~Michalove, M.~Schmittle, C.~Summers,
  M.~Rockett, J.~R. Smith, S.~Chouhury, C.~Mavrogiannis, and F.~Sadeghi,
  ``{MuSHR}: A low-cost, open-source robotic racecar for education and
  research,'' {\em CoRR}, vol.~abs/1908.08031, 2019.

\bibitem{gazebo}
Open-Source-Robotics-Foundation, ``A {3D} multi-robot simulator with
  dynamics.'' \url{http://gazebosim.org/}.
\newblock Accessed: 2022-03-01.

\bibitem{racecarMIT2022}
\relax RACECAR~Team, ``The racecar project.'' \url{https://racecar.mit.edu},
  2022.

\bibitem{racecarMIT-github2022}
\relax RACECAR~Team, ``Github repo, racecar project.''
  \url{https://github.com/mit-racecar}, 2022.

\bibitem{catPlatform2022}
J.~Tse, ``The cat platform.''
  \url{https://www.jontse.com/portfolio/autonomous_rc_car.html}, 2022.

\bibitem{donkeyCar2022}
\relax Donkey Car~Community, ``The donkey car.''
  \url{https://docs.donkeycar.com}, 2022.

\bibitem{unityGaming}
Unity3D, ``{Main Website}.'' \url{https://unity3d.com/}, 2016.
\newblock Accessed: 2021-11-23.

\bibitem{physxNVIDIA}
\relax{NVIDIA}, ``{PhysX} simulation engine.'' Available online at
  \url{http://developer.nvidia.com/object/physx.html}, 2019.

\bibitem{chronoOverview2016}
A.~Tasora, R.~Serban, H.~Mazhar, A.~Pazouki, D.~Melanz, J.~Fleischmann,
  M.~Taylor, H.~Sugiyama, and D.~Negrut, ``{Chrono}: An open source
  multi-physics dynamics engine,'' in {\em {High Performance Computing in
  Science and Engineering -- Lecture Notes in Computer Science}} (T.~Kozubek,
  ed.), pp.~19--49, Springer International Publishing, 2016.

\bibitem{pyChronoCondaWebSite}
\relax {Project Chrono} Development~Team, ``{PyChrono}: A python wrapper for
  the chrono multi-physics library.''
  \url{https://anaconda.org/projectchrono/pychrono}.
\newblock Accessed: 2021-11-19.

\bibitem{ChronoVehicle2019}
R.~Serban, M.~Taylor, D.~Negrut, and A.~Tasora, ``{Chrono::Vehicle}
  template-based ground vehicle modeling and simulation,'' {\em Intl. J. Veh.
  Performance}, vol.~5, no.~1, pp.~18--39, 2019.

\bibitem{chronoSCM2019}
A.~Tasora, D.~Mangoni, D.~Negrut, R.~Serban, and P.~Jayakumar, ``Deformable
  soil with adaptive level of detail for tracked and wheeled vehicles,'' {\em
  International Journal of Vehicle Performance}, vol.~5, no.~1, pp.~60--76,
  2019.

\bibitem{weiTracCtrl2022}
W.~Hu, Z.~Zhou, S.~Chandler, D.~Apostolopoulos, K.~Kamrin, R.~Serban, and
  D.~Negrut, ``Traction control design for off-road mobility using an
  {SPH}-{DAE} co-simulation framework,'' {\em Multibody System Dynamics},
  vol.~\text{ }, no.~https://doi.org/10.1007/s11044-022-09815-2, 2022.

\bibitem{asherSensorSimulation2021}
A.~Elmquist, R.~Serban, and D.~Negrut, ``A sensor simulation framework for
  training and testing robots and autonomous vehicles,'' {\em Journal of
  Autonomous Vehicles and Systems}, vol.~1, no.~2, p.~021001, 2021.

\bibitem{synchrono2020}
J.~Taves, A.~Elmquist, A.~Young, R.~Serban, and D.~Negrut, ``Synchrono: A
  scalable, physics-based simulation platform for testing groups of autonomous
  vehicles and/or robots,'' in {\em {Proceedings of 2020 International
  Conference on Intelligent Robots and Systems (IROS) -- Las Vegas, USA}},
  2020.

\bibitem{atk-pypi}
\relax {UW-Madison Simulation Based Engineering Laboratory}, ``{PyPI:
  Autonomy-ToolKit}.'' \url{https://pypi.org/project/autonomy-toolkit/}, 2022.

\bibitem{atk-art2022}
\relax {UW-Madison Simulation Based Engineering Laboratory}, ``{Autonomy
  Toolkit}.'' \url{http://projects.sbel.org/autonomy-toolkit/}, 2022.

\bibitem{ren2016faster}
S.~Ren, K.~He, R.~Girshick, and J.~Sun, ``Faster {R-CNN}: Towards real-time
  object detection with region proposal networks,'' {\em {IEEE} transactions on
  pattern analysis and machine intelligence}, vol.~39, no.~6, pp.~1137--1149,
  2016.

\bibitem{mobileNetV3}
A.~Howard, M.~Sandler, B.~Chen, W.~Wang, L.~Chen, M.~Tan, G.~Chu, V.~Vasudevan,
  Y.~Zhu, R.~Pang, H.~Adam, and Q.~Le, ``Searching for mobilenetv3,'' in {\em
  2019 IEEE/CVF International Conference on Computer Vision (ICCV)}, (Los
  Alamitos, CA, USA), pp.~1314--1324, IEEE Computer Society, nov 2019.

\bibitem{art-iros-video}
\relax {UW-Madison Simulation Based Engineering Laboratory}, ``{ART Supporting
  Video}.'' \url{https://uwmadison.box.com/s/iypwo22vj26yqjqcprhc65trbe5vp8vz},
  2022.

\end{thebibliography}
\def\cprime{$'$}

\end{document}